\pdfoutput=1

\documentclass[11pt]{article}

\usepackage[final]{acl}
\usepackage{times}
\usepackage{latexsym}

\usepackage[T1]{fontenc}

\usepackage[utf8]{inputenc}

\usepackage{microtype}

\usepackage{inconsolata}

\usepackage{amsmath,amsfonts,bm}

\def\eqref#1{equation~\ref{#1}}

\def\1{\bm{1}}

\DeclareMathAlphabet{\mathsfit}{\encodingdefault}{\sfdefault}{m}{sl}
\SetMathAlphabet{\mathsfit}{bold}{\encodingdefault}{\sfdefault}{bx}{n}

\usepackage{graphicx}  
\usepackage{float}
\usepackage{wrapfig} 
\usepackage{hyperref}
\usepackage{xspace}
\usepackage{url}
\usepackage{booktabs}
\usepackage{MnSymbol,bbding,pifont}
\usepackage{colortbl}
\definecolor{LightCyan}{rgb}{0.88,1,1}
\usepackage{enumitem}
\usepackage{multirow}
\usepackage{xcolor}

\usepackage{color-edits}
\addauthor{gn}{magenta}
\addauthor{ys}{orange}
\definecolor{browse}{rgb}{0.68, 0.87, 0.56}
\definecolor{api}{rgb}{0.50, 0.80, 0.73}
\definecolor{hybrid}{rgb}{0.30, 0.72, 0.80}

\newcommand{\hlbrowse}[1]{\colorbox{browse!100}{#1}}
\newcommand{\hlapi}[1]{\colorbox{api!100}{#1}}
\newcommand{\hlhybrid}[1]{\colorbox{hybrid!100}{#1}}
\usepackage{tcolorbox}

\newtcolorbox{promptbox}[2][Prompt]{
colback=black!5!white,
arc=5pt, 
boxrule=0.5pt,
fonttitle=\bfseries,
title=#1, 
before upper={\small}, fontupper=\fontfamily{ptm}\selectfont,
colframe=#2,
}

\title{Beyond Browsing: API-Based Web Agents}

\author{Yueqi Song, Frank Xu, Shuyan Zhou, Graham Neubig \\
\texttt{\{yueqis,gneubig\}@cs.cmu.edu} \\
Carnegie Mellon University \\ \\
\url{https://yueqis.github.io/API-Based-Agent/}
}

\newcommand{\todo}[1]{\textcolor{green}{[]}}

\begin{document}
\maketitle

\begin{abstract}
Web browsers are a portal to the internet, where much of human activity is undertaken. Thus, there has been significant research work in AI agents that interact with the internet through web browsing.
However, there is also another interface designed specifically for machine interaction with online content: application programming interfaces (APIs).
In this paper we ask -- \emph{what if we were to take tasks traditionally tackled by Browsing Agents, and give AI agents access to APIs}?
To do so, we propose two varieties of agents: (1) an API-calling agent that attempts to perform online tasks through APIs only, similar to traditional coding agents, and (2) a Hybrid Agent that can interact with online data through both web browsing and APIs.
In experiments on WebArena, a widely-used and realistic benchmark for web navigation tasks, we find that API-Based Agents outperform web Browsing Agents.
Hybrid Agents out-perform both others nearly uniformly across tasks, resulting in a more than 24.0\% absolute improvement over web browsing alone, achieving a success rate of 38.9\%, the SOTA performance among task-agnostic agents.
These results strongly suggest that when APIs are available, they present an attractive alternative to relying on web browsing alone.
\end{abstract}

\section{Introduction}
Web agents use browsers as an interface to facilitate humans in performing daily tasks such as online shopping, online planning, trip planning, and other work-related tasks \citep{zheran2018reinforcement,li-etal-2020-mapping,rawles2023androidinthewild,patil2023gorilla,pan2024autonomous,chen2024internet,huang2024positionpaperagentai,durante2024interactiveagentfoundationmodel}.
Existing web agents typically operate within the space of graphical user interfaces (GUI) \citep{yang2023appagent, zhou2023webarena,zheng2024seeact}, using action spaces that simulate human-like keyboard and mouse operations, such as clicking and typing.
To observe webpages, common approaches include using accessibility trees, a simplified version of the HTML DOM tree, as input to text-based models \citep{zhou2023webarena,drouin2024workarena}, or multi-modal, screenshot-based models \citep{koh-etal-2024-visualwebarena, OSWorld, you2024ferretuigroundedmobileui, hong2023cogagentvisuallanguagemodel}.
However, regardless of the interaction method with websites, there is no getting around the fact that these sites were originally designed for humans, and may not be the ideal interface for machines.
\begin{figure*}[!h]
    \centering
    \includegraphics[width=\textwidth]{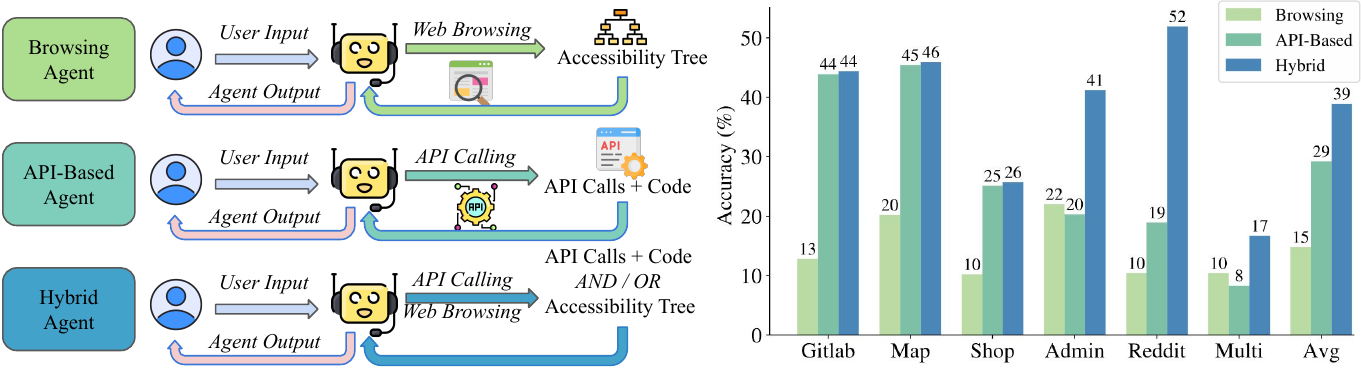}
    \caption{The \hlbrowse{Browsing Agent} performs tasks through browsing only, utilizing the accessibility tree to interact with webpages, achieving an average performance of 14.8\% on WebArena. Without reliance on web browsing, the \hlapi{API-Based Agent} performs tasks by making API calls and generating code without relying on web browsing, achieving an average accuracy of 29.2\%. Dynamically interleaving web browsing and API calling, the \hlhybrid{Hybrid Agent} executes either API calls or browsing actions, or combining both, achieving performance of 38.9\%.}
    \label{fig:main}
\end{figure*}

Notably, there is another interface designed specifically for machine interaction with the web: application programming interfaces (APIs) \citep{chan2024visibility}. APIs allow machines to communicate directly with backends of web services \citep{branavan-etal-2009-reinforcement}, sending and receiving data in machine-friendly formats such as JSON or XML \citep{meng2018application,xu-etal-2021-grounding,appworld-acl24}. Nonetheless, whether AI agents can effectively use APIs to tackle real-world online tasks, and the conditions under which this is possible, remain unstudied.
In this work, we explore methods for tackling tasks normally framed as web-navigation tasks with an expanded action space to interact with APIs.
To do so, we develop new \emph{API-Based Agents} that directly interact with web services via API calls. This method bypasses the need to interact with web GUIs.

However, not all websites have extensive API support, in which case web browsing actions may still be required. 
To overcome this limitation, we propose a \emph{hybrid} approach that combines API-Based Agents with Browsing Agents, as depicted in Figure \ref{fig:main}. 
Rather than choosing between API calling and web browsing at the task level, our Hybrid Agent is capable of dynamically \emph{interleaving} both actions. We found that agents benefit from the flexibility of this hybrid model. When APIs are available and well-documented, the agent can directly interact with the web services. For websites with limited API support, the agent seamlessly interleaves API calling and browsing, combining the power of both to complete each task.

We evaluated our API-Based and Hybrid Agents on WebArena, a benchmark for real-world web tasks \citep{zhou2023webarena}, and the results are shown in Figure \ref{fig:main}. Our experiments revealed three key findings: (1) The API-Based Agent outperforms the Browsing Agent on WebArena by around 15\% on average. 
(2) The API-Based Agent yields a higher success rate on websites with good API support (e.g., Gitlab) compared to those with limited API support (e.g., Reddit). 
This result underscores the importance of developing comprehensive API support for more accurate and efficient web task automation in the future.
(3) The Hybrid Agent outperforms solely Browsing and solely API-Based Agents, further improving accuracy by 10\% compared to the API-Based Agent.
By dynamically interleaving approaches, the Hybrid Agent is able to provide more consistent and reliable outcomes.

In sum, our results suggest that allowing agents to interact with APIs, interfaces designed specifically for machines is often preferable or at least complementary to direct interaction with graphical interfaces designed for humans.

\section{Background: Web Browsing}

\subsection{The Web Browsing Task}
\label{subsec:benchmark}

\begin{figure*}[t]
    \centering
    \includegraphics[width=\textwidth]{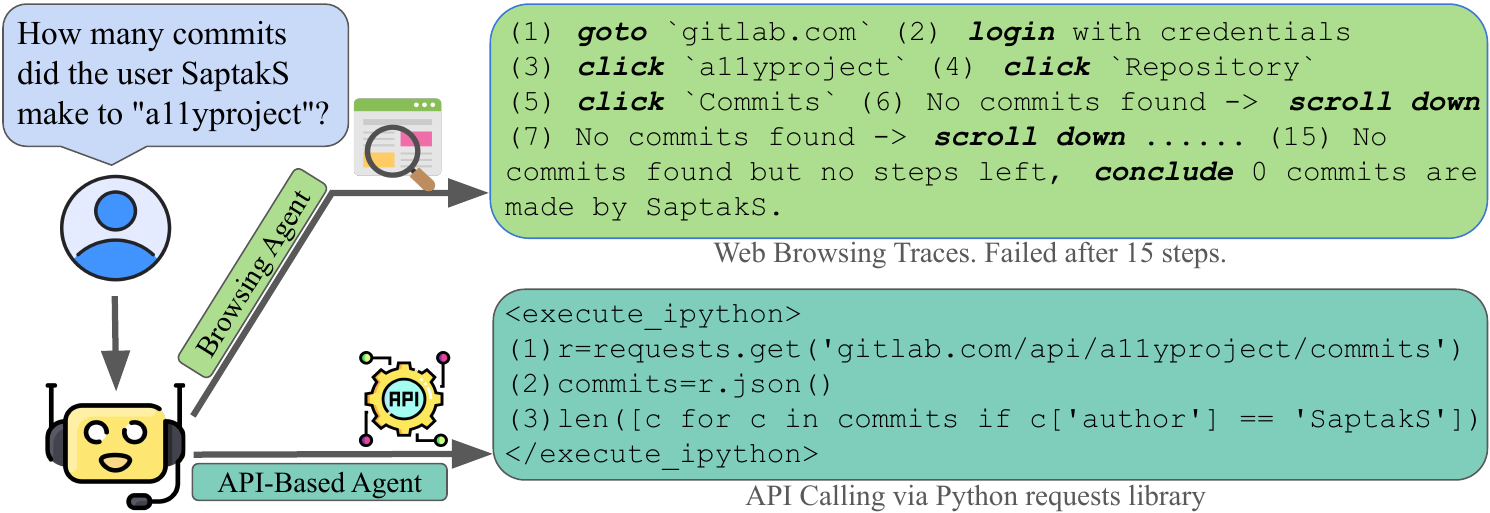} 
    \caption{The \hlapi{API-Based Agent} often solves problems in fewer steps than the \hlbrowse{Browsing Agent}. In this task, web browsing failed to solve the intent ``find the number of commits the user \textit{SaptakS} made to the repo \textit{a11yproject}'' after 15 steps, while the \hlapi{API-Based Agent} successfully completed the task with only three lines of code.}
    \label{fig:example}
\end{figure*}

Various benchmarks have been developed to evaluate web Browsing Agents. MiniWoB (Miniature World of Bits) is an early benchmark that provides simple web-based tasks such as clicking links or typing into forms, but it remains limited in complexity and realism \citep{pmlr-v70-shi17a}. Mind2Web scales up these tasks, introducing more complex interactions across websites, but it primarily focuses on basic web operations \citep{deng2023mind2web}. WebArena \citep{zhou2023webarena} advances web browsing benchmarks by creating reproducible sandboxes of various websites, such as managing repositories, posting online, performing online shopping, and planning trips using map services, while VisualWebArena extends WebArena to the vision modality \citep{koh-etal-2024-visualwebarena}. 

In this paper, we focus on WebArena tasks, which simulate real-world scenarios to evaluate an agent's ability to complete diverse web-based activities.%
\footnote{Notably, upon investigation of  VisualWebArena we found that APIs for handling images were relatively limited, and hence we chose to experiment on text-only tasks in this paper.}
WebArena tasks include interacting with platforms like Gitlab (to manage projects and repositories), Reddit (to browse and post content), e-commerce websites (for shopping), and mapping services (for trip planning) \citep{zhou2023webarena}. Task success is evaluated in three ways: (1) if the task requires producing specific outputs, agents' responses are checked for correctness; (2) for tasks involving changes to a website's state (e.g., adding items to shopping carts), success is verified by whether the state has changed as expected, such as ensuring the correct items have been added to the cart; and (3) if the task involves navigation, success is determined by whether the agent reaches the correct URL displaying the desired content.

\subsection{A Baseline Web Browsing Agent}
\label{sec:baseline}

While there are a wide variety of agents proposed for such web navigation tasks, in this work we build upon the WebArena baseline agent \citep{zhou2023webarena}, which operates purely through web interaction by leveraging the accessibility tree\footnote{\url{https://developer.mozilla.org/en-US/docs/Glossary/Accessibility_tree}}, a structure that exposes interactive elements like buttons, input fields, and hyperlinks \citep{yao2023webshopscalablerealworldweb,gu2024middlewarellmstoolsinstrumental}. Each element of the accessibility tree is characterized by its functionality such as a hyperlink, its content, and specific web attributes \citep{liu2024visualagentbench,he-etal-2024-webvoyager,lù2024weblinxrealworldwebsitenavigation}.
This exposes webpage elements in a hierarchical structure that is easy for agents to navigate \citep{samuel2024personagymevaluatingpersonaagents,10.1007/978-3-031-20074-8_18}. 

Agents based on this framework utilize an action space that simulates human browsing behavior, incorporating actions such as simulated clicks, form input, and navigation between pages \citep{liu2023agentbench,song-etal-2024-trial,gur2024a}. Importantly, these agents maintain a comprehensive history of their previous actions, allowing them to contextualize their decision-making in past actions.

While agents utilizing this method can navigate arbitrary webpages and often perform well on simple layouts, challenges arise with the complexity of GUIs. Many large language models (LLMs) are not familiar with accessibility trees, which leads to difficulties in completing tasks that require numerous or complex interactions, resulting in lower accuracies \citep{liu2024visualwebbenchfarmultimodalllms,deng2023mind2web,fu2024autoguide}. These methods also struggle with content that need to be dynamically loaded or contents not immediately visible within the tree \citep{abramovich2024enigmaenhancedinteractivegenerative,chen-etal-2024-agent,lutz2024wilbur}.

To give a motivating example, in Figure \ref{fig:example}, we demonstrate a task where agents need to determine the number of commits made by the user \textit{SaptakS} in a repository named \textit{a11yproject}. For each task, agents are given a fixed number of steps within which to complete the task. Using a traditional browsing approach, the agent follows a complex trajectory, starting with logging in, navigating to the correct project, accessing the repository, and finally attempting to view the list of commits. However, due to the large number of commits made by other users, the commits by \textit{SaptakS} are located much further down on the webpage, requiring agents to scroll down many times. As a result, despite completing 15 steps, the Browsing Agent is unable to retrieve the required information.

\section{From Web Browsing to API Calling}

\begin{figure*}[t]
    \centering
    \includegraphics[width=15cm]{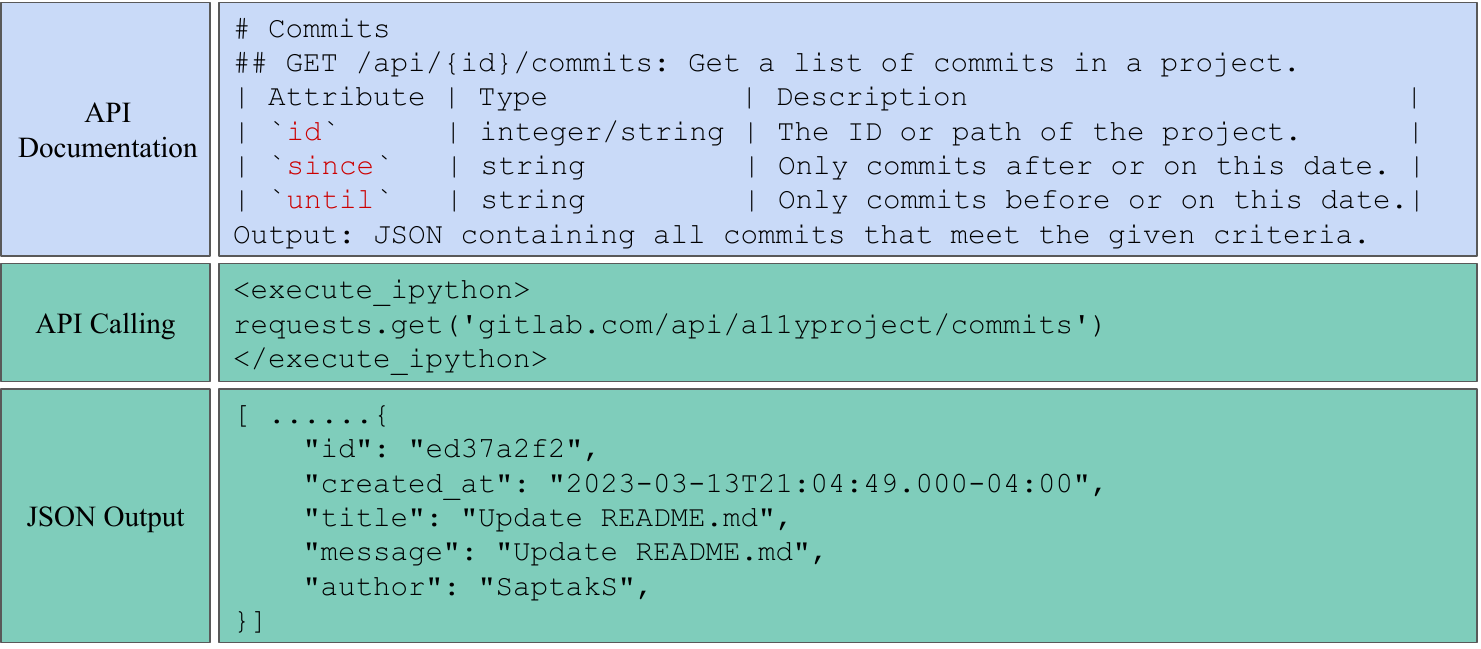} 
    \caption{An example of API documentation showing how to get commits of a project, the API call using a Python script to retrieve commits from a project repository, and the resulting JSON response.}
    \label{fig:api}
\end{figure*}

In contrast, API calling allows machines to directly communicate with web services, reducing operational complexity.
In this section, we explore an API-based approach when performing web tasks.

\subsection{APIs and API Documentation}


For websites that offer API support, pre-defined endpoints can be utilized to perform tasks efficiently. 
These APIs, following standardized protocols like REST\footnote{\url{https://en.wikipedia.org/wiki/REST}}, allow interaction with web services through sending HTTP requests (e.g., \texttt{GET}, \texttt{POST}, \texttt{PUT}) and receiving structured data such as JSON objects\footnote{\url{https://www.json.org/json-en.html}} as responses. Websites often provide official documentation for the APIs, which can give guidance on how to utilize the APIs. Some documentation is provided as plain text, some in README \footnote{\url{https://en.wikipedia.org/wiki/README}} format, and some in OpenAPI YAML\footnote{\url{https://yaml.org/}} format. Figure \ref{fig:api} shows an example of the Gitlab README documentation of \texttt{GET /api/\{id\}/commits}, documenting it's functionality, input arguments, and output types. For example, to retrieve all commits to \texttt{a11yproject}, one could use the Python \texttt{requests} library, by calling \texttt{requests.get}``\texttt{gitlab.com/api/a11yproject/}\linebreak\texttt{commits}''\texttt{)}. This returns a JSON list containing all the commits to this repo, as shown in Figure \ref{fig:api}.




\subsection{Obtaining APIs for Agents}
One important design decision is how to obtain APIs for agents to use. The way agents interact with APIs depends heavily on the availability of APIs and quality of API documentation. In this work, we acquired APIs by manually looking up official API documentation on a website, although this process could potentially be automated in the future.
We classify the availability of APIs according to the following three scenarios:


\paragraph{Sufficient APIs and Documentation} Many websites provide comprehensive API support and well-documented API documentation in YAML or README format.
In this case, simply use the APIs/documentation as-is.
Figure \ref{fig:api} depicts an example of API documentation.

\paragraph{Sufficient APIs, Insufficient Documentation} There are some challenging situations where APIs exist but good documentation is not officially available. In such cases, additional steps may be required to obtain a list of accessible APIs. In this case, we inspected the frontend or backend code of the website to extract undocumented API calls that can still be utilized by the agent. Then, based on the implementation of APIs of the website, leverage an LLM (GPT-4o\footnote{\url{https://openai.com/index/hello-gpt-4o/}}) to generate these YAML or README files.
By prompting GPT-4o with the relevant implementation details of the APIs (for example, the implementation files of the APIs or example traces of API calls), we generate comprehensive documentation, including input parameters, expected outputs, and example API calls.

\paragraph{Insufficient APIs} In the more challenging cases, where only minimal APIs are available, it may be necessary to create new APIs. These custom APIs allow agents to perform tasks that otherwise would require manual web browsing steps. In our case, this was necessary for 1 of 5 websites in the WebArena benchmark that we utilized, such as creating Reddit APIs discussed in Section \ref{sec:quality}.

\subsection{Using APIs in Agents}
\label{subsec:prompt}

Once we have the APIs and documentation, we then need to provide methods to utilize them in agents.
We utilize two different methods based on the size of the API documentation.

\paragraph{One-Stage Documentation for Small API Sets} For websites with smaller numbers of APIs\footnote{We use a threshold of 100 APIs, but this could be adjusted depending on the supported language model context size.}, we directly incorporate the full documentation into the prompt provided to the agent. 
This approach of directly feeding the full documentation worked well for websites with a limited number of API endpoints, as it allowed the agent to have immediate access to all the necessary information without the need for a more complex retrieval mechanism.

\paragraph{Two-Stage Documentation Retrieval for Large API Sets} For websites with more APIs, providing the full documentation in the prompt is impractical due to size limitation of agent inputs. To address this, we use a two-stage documentation retrieval process, allowing access to only the needed information to keep the initial prompt concise.

In the first stage, the user prompt provides a task description, with a list of all available APIs along with a brief description of each. For example, \texttt{\{}``\texttt{GET /api/\{id\}/commits}'': ``\texttt{List commits in a project}''\texttt{\}}. This initial summary helps in understanding the scope of all the available APIs while staying within the prompt size constraints.

In the second stage, if the model determines that it needs detailed information about one or more specific API endpoints, it can use a tool named \texttt{get\_api\_documentation}, which maintains a dictionary that maps each API to its documentation respectively. The dictionary is generated using Python pattern match to retrieve substrings related to each endpoints. This tool is able to search the dictionary and retrieve the full README or YAML documentation for any given endpoint with the endpoint's identifier. The resulting documentation might include the input parameters, output formats, and examples of how to interact with the endpoint. For example, to retrieve the documentation for the API \texttt{GET /api/{id}/commits}, the agent would call \texttt{get\_api\_documentation(``GET /api/{id}/commits'')}. An example returned API documentation is the documentation in Figure \ref{fig:api}. 

This retrieval method allows the agent to make flexible and informed decisions to perform tasks. If the agent finds that an API does not meet its needs or if it encounters an error, it can easily retrieve the documentation for a different API by calling the tool again. This dynamic approach promotes adaptability and minimizes the risk of incorrect API usage when the number of APIs available is large.
The prompt can be found in Appendix \ref{appendix:api-prompt}.

\section{Hybrid Browsing+API Calling Agents}

\label{sec:hybrid}

We have proposed API-based methods for handling web tasks, but the question arises: given the benefits of API calling, should we discard browsing altogether? 
The most obvious bottleneck is that not all websites offer good API support. 
Some platforms offer limited or poorly documented APIs (e.g. no API for shopping on Amazon\footnote{\url{https://www.amazon.com}}), forcing agents to rely on browsing to complete tasks. 

To deal with these situations, we propose a hybrid methods that integrates both browsing and API calling, and developed a Hybrid Agent capable of dynamically interleaving API calls and web browsing based on task requirements and the available resources.
Specifically, for each task, the agent is given the fixed step budget within which it has to finish the task.
\textit{In each step of a task}, the agent could either (1) communicate with humans in natural language to ask
for clarification, or 2) generate and executes Python code which could include performing API calling, or 3) performs web browsing actions. The Hybrid Agent could choose freely among these options, depending on the agent's confidence in which method is the best for each step.

Ideally, for websites with good API support, the Hybrid Agent can utilize well-documented APIs to perform tasks more efficiently than it could through only browsing; for websites with limited API support or poor documentation, the Hybrid Agent could rely more on browsing. We find that enabling it to interleave API calling and web browsing boosts task performance (see Section \ref{sec:results}).

\paragraph{Prompt Construction} The Hybrid Agent's prompt construction extends upon the API-Based Agent by incorporating both API and web-browsing documentation. Similar to the API-Based Agent, the Hybrid Agent is provided with a description of available API calls as discussed in Section \ref{subsec:prompt}. In addition, the Hybrid Agent receives a detailed specification of the web-browsing actions, which mirrors the information given to the Browsing Agent described in Section \ref{sec:baseline}, including a breakdown of all potential browser interactions. It also maintains a history of all its prior steps such that the agent could make more informed actions.
The prompt can be found in Appendix \ref{appendix:hybrid-prompt}.

\section{Experimental Setup}

\label{experiment}

\subsection{Dataset Description}
We utilized WebArena \citep{zhou2023webarena} as the primary evaluation benchmark. WebArena is a comprehensive benchmark designed for real-world web tasks, providing a diverse set of websites that simulate various online interactions, allowing comprehensive evaluation of agents' abilities to handle both API calling and web browsing across varied web settings. 
WebArena mainly includes five websites, each with various intents representing different tasks: 
Gitlab, Map, Shopping, Shopping Admin, Reddit, and Multi-Site tasks. A detailed descriptions of the tasks is in Appendix \ref{appendix:webarena}.

\subsection{API Statistics for WebArena Sites}
\label{sec:api-stats}
The API support for WebArena websites can be categorized into three levels: good, medium, and poor. APIs' availability, functionality, and documentation, as described in Table \ref{tab:quality}, play a crucial role in the efficiency and flexibility of our agents\footnote{See Appendix \ref{appendix:apis} for where to find the WebArena APIs.}.

\begin{table}[!h]
\centering
\resizebox{\linewidth}{!}{
\begin{tabular}{l|llllll}
\toprule
\textbf{Sites} & \textbf{Gitlab} & \textbf{Map} & \textbf{Shop} & \textbf{Admin} & \textbf{Reddit} \\ \midrule
\# APIs & 988 & 53 & 556 & 556 & 31$^\ddagger$ \\ 
Quality & Good & Good & Fair & Fair & Poor\\
\bottomrule
\end{tabular}}
\vspace{-3mm}
\caption{Number of endpoints, and quality of API and documentation for WebArena websites. $^\ddagger$ See Section \ref{sec:quality} for more discussions.}
\label{tab:quality}
\vspace{-3mm}
\end{table}

\subsubsection{Good API Support}

\paragraph{Gitlab}

Gitlab supports 988 endpoints, which offer extensive coverage across a wide range of functionalities, including repositories, commits, and users. This comprehensive API support allows for effective interaction in most WebArena tasks, making Gitlab one of the best-supported platforms in terms of API availability.

\noindent\textbf{Map} The Map website offers 53 endpoints. 
Despite the smaller number of endpoints, the APIs available are well-documented and cover most of the essential WebArena use cases.

\subsubsection{Medium API Support}

\paragraph{Shopping and Shopping Admin}

The Shopping and Shopping Admin websites share a common set of 556 APIs, which provide a reasonable level of support for common shopping tasks. However, some features, such as adding items to wish lists, are absent, and thus these tasks must be handled via browsing. Despite this, the documentation is fairly detailed. Overall, API calling is a solid, though not exhaustive, solution for handling shopping tasks.

\subsubsection{Poor API Support}

\paragraph{Reddit} The WebArena Reddit is a self-hosted limited clone of the actual Reddit\footnote{See Appendix \ref{appendix:apis} for more explanations.} with only 31 endpoints. It offers minimal API support and no documentation, making it the least API-friendly site in WebArena. Many critical functionalities such as searching posts are missing, significantly hampering task execution on Reddit, highlighting the need for a hybrid browsing+API approach.


\subsection{API Implementation Details}

We follow the methodologies discussed in Section \ref{subsec:prompt} to provide APIs to agents. Appendix \ref{appendix:apis} contains the sources of the public API documentations.

\subsubsection{One-Stage Documentation for Small API Sets} For websites with fewer than 100 API endpoints, namely the Map and Reddit websites, we directly provide the full documentation to the agent. 

\noindent\textbf{Map } The README documentation was inputted directly from the public API documentation. 


\noindent\textbf{Reddit } Since there was no pre-existing documentation for the APIs, we leveraged GPT-4o\footnote{\url{https://openai.com/index/hello-gpt-4o/}} itself to generate these README files. By prompting GPT-4o with a file containing all implementations of the API endpoints, we generated a README documentation, including input parameters, expected outputs, and example API calls. 

\subsubsection{Two-Stage Documentation Retrieval for Large API Sets} For websites with more than 100 endpoints, namely GitLab, Shopping, and Shopping Admin, we employ a two-stage documentation retrieval process.

We obtained Gitlab README documentations from the official website. For Shopping and Shopping Admin, the documentation is provided as OpenAPI specification, structured in YAML format.

\subsection{Evaluation Framework}
\begin{table*}[!h]
\centering
\resizebox{\textwidth}{!}{
\begin{tabular}{l|llllll|l}
\toprule
\textbf{Agents} & \textbf{Gitlab} & \textbf{Map} & \textbf{Shopping} & \textbf{Admin} & \textbf{Reddit} & \textbf{Multi} & \textbf{AVG.} \\ \midrule
WebArena Base \citep{zhou2023webarena} & 15.0 & 15.6 & 13.9 & 10.4 & 6.6 & 8.3 & 12.3 \\
AutoEval \citep{pan2024autonomous} & 25.0 & 27.5 & 39.6 & 20.9 & 20.8 & 16.7 & 26.9\\
AWM \citep{wang2024agent} & 35.0 & 42.2 & 32.1 & 29.1 & 54.7 & \textbf{18.8} & 35.5 \\ 
SteP \citep{sodhi2024step}$^\dagger$ & 32.2 & 31.2 & \textbf{50.8} & 23.6 & \textbf{57.5} & 10.4 & 36.5 \\ \midrule
Browsing Agent & 12.8 & 20.2 & 10.2 & 22.0 & 10.4 & 10.4 & 14.8\\
API-Based Agent & 43.9 & 45.4 & 25.1 & 20.3 & 18.9$^\ddagger$ & 8.3 &  29.2   \\
Hybrid Agent & \textbf{44.4} & \textbf{45.9} & 25.7 & \textbf{41.2} & 51.9$^\ddagger$ & 16.7 &  \textbf{38.9}
 \\ \bottomrule
\end{tabular}}
\caption{Agents' performances on WebArena. $^\dagger$Note that SteP uses prompts inspired specifically by WebArena tasks, while other agents are task-agnostic. Our Hybrid Agent achieve the highest accuracy among these agents. $^\ddagger$ We report these results using a set of APIs introduced by the authors to mimic the official Reddit website, constructed task-agnostically without access to WebArena tasks. See Section \ref{sec:quality} for more discussions.
}
\label{main-results}
\vspace{-2mm}
\end{table*}

We employed OpenHands as our evaluation framework to facilitate the development and testing of our agents \citep{wang2024opendevin}.
OpenHands is an open-source platform designed for creating and evaluating AI agents that interact with both software and web environments, making it an appropriate infrastructure for our proposed methods.
The OpenHands architecture supports various interfaces for agents to interact with. Moreover, this framework allows agents to keep a detailed record of past actions in the prompt, enabling agents to execute actions in a way that is consistent with earlier steps. For coding tasks, it implements an agent based on CodeAct \citep{wang2024executable} that incorporates a sandboxed bash operating system and Jupyter IPython\footnote{\url{https://ipython.org}} environments, enabling Python code execution.
Additionally, it includes a BrowsingAgent Browsing Agent that focuses solely on web navigation. This agent operates within a Chromium web browser powered by Playwright\footnote{\url{https://playwright.dev/}}, utilizing a comprehensive set of browser actions defined by BrowserGym \citep{workarena2024}. 
However, while the Browsing Agent can browse websites, and the CodeActAgent make API calls and execute code, there is not an agent that can natively do both.
Given this base, we developed two varieties of agents for API-based solving of web tasks.

\paragraph{API-Based Agent}
Our API-Based Agent essentially uses the CodeAct architecture \citep{wang2024executable}.
In addition to the basic CodeAct framework, we tailor the agent for API calling by adding specialized instructions and examples that guide its understanding and using of APIs. At each step, the agent could utilize all previous actions to make informed selection of actions. The prompt of the API-Based Agent is included in the Appendix \ref{appendix:api-prompt}.

\paragraph{Hybrid Browsing/API Calling Agent}
In addition to the API-Based Agent, we developed a Hybrid Agent that integrates Chromium web browsing functionalities powered by Playwright into the existing API-Based Agent framework.
This Hybrid Agent is provided the prompt describing both the APIs and the browsing actions, allowing for free transitions between API calling and web browsing. At each step, the agent can utilize the current state of the browser, all previous actions taken by the agent, and the results of those actions to determine the next course of action.
The prompt of the Hybrid Agent is included in the Appendix \ref{appendix:hybrid-prompt}.

For the Browsing, API-Based, and Hybrid Agents, we utilized GPT-4o as the base LLM. However, this could be easily changed to other LLMs.

\section{Results}
\label{sec:results}

\subsection{Main Results}


The main results of our evaluation, as summarized in Table \ref{main-results}, demonstrate the performance of three different agents across WebArena websites.

The API-Based Agent consistently achieved higher scores on most websites compared to the Browsing Agent. This agent's strong performance is attributed to its specialized design for API calling, enabling it to efficiently interact with websites and complete tasks with no reliance on browsing. 

In contrast, the Browsing Agent, designed solely for navigating web interfaces, demonstrated significantly lower performance across most domains. It achieved its best scores on Shopping Admin and Map, but struggled more on the other websites.

The Hybrid Agent, integrating both API calling and web browsing, outperformed the Browsing and API-Based Agents on all websites. The agent's ability to interleave API calling and web browsing proved beneficial. 
API calling delivers high performance for web tasks when well-supported APIs are available,
while web browsing serves as a backup when API endpoints are unavailable or incomplete. Even if the website provides comprehensive APIs, there might be corner cases where APIs are not supportive. Thus, relying on web browsing is still needed for tasks that would otherwise fail through API-only interactions.
Table \ref{action_percent} show the frequency of each action type of the Hybrid Agent: it chooses to do both Browsing and API in 77.7\% of WebArena tasks, and it shows higher accuracy when choosing API only and API+browsing. More detailed analysis on action types, steps and cost and case studies are in Appendix \ref{appendix:analysis} and \ref{appendix:case}. 

\begin{table}[!h]
\centering
\resizebox{\linewidth}{!}{
\begin{tabular}{l|r|r}
\toprule
\textbf{Actions} & \textbf{Frequency (\%)} & \textbf{Accuracy (\%)} \\ \midrule
Browsing only & 14.3 & 25.0\\
API only & 8.0 & 40.0\\
Browsing+API & 77.7 & 42.0\\ 
\bottomrule
\end{tabular}}
\caption{The left column specifies the type of action taken by the Hybrid Agent; the middle column shows the percentage of actions selected among all WebArena tasks; and the right column indicates the accuracy of the Hybrid Agent for each action type.}
\label{action_percent}
\end{table}

Overall, the results indicate that the Hybrid Agent is the most effective for handling diverse tasks in WebArena, particularly in environments that require a blend of API and browsing actions. The API-Based Agent excels in tasks that are primarily API-driven, while the Browsing Agent is more suitable for simple navigation tasks but lacks the versatility needed for more complex scenarios. 






\subsection{Does API Quality Matter?}
\label{sec:quality}



Yes, API quality does significantly impact the performance of agents. High quality APIs provide comprehensive and well-documented endpoints that enable agents to interact accurately and efficiently with websites. With comprehensive API support, the API-Based Agent is able to tackle more tasks through API calling, while the Hybrid Agent rely less on browsing; on the other hand, clear and detailed documentation allows agents to use APIs effectively, ensuring that requests are accurate, and minimizing potential errors in task execution. For example, Gitlab and Map, with the best API support as mentioned in Section \ref{sec:api-stats}, demonstrate highest task completion accuracies among websites by the API-Based and Hybrid Agent.

Conversely, low-quality APIs, characterized by incomplete functionality or ambiguous documentation, can significantly degrade performance. In such cases, the absence of necessary endpoints may prevent the API-Based Agent from completing tasks. Moreover, poorly documented APIs can result in misusing parameters and headers, further reducing the effectiveness of the agent. This highlights the importance for websites to maintain comprehensive and well-documented API support.

\begin{table}[!h] 
\centering
\small
\begin{tabular}{ccc}
\toprule
\textbf{Number of Endpoints} & 18 & 31\\
\midrule
\textbf{Accuracy on Reddit} & 9.4\% & 18.9\% \\
\bottomrule
\end{tabular}
\caption{{Change in performance of the API-Based Agent on Reddit upon incorporating new APIs.}}
\label{tab:change}
\vspace{-5mm}
\end{table}
An illustrative example of this is the case of Reddit, where the initial performance of the API-Based Agent was suboptimal due to limited API availability. 
As depicted in Table \ref{tab:change}, initially, Reddit offered only 18 APIs, lacking the major functionality that common online forums have, such as post voting. 
Recognizing this limitation, we manually introduced 13 additional APIs including one API on post voting, with our best effort trying to mimic the official Reddit website. This results in a marked improvement in the API-Based Agent's performance, underscoring the direct correlation between the availability of high-quality APIs and the average performance of the API-Based Agent.


\begin{figure}[!h]
    \centering
    \includegraphics[width=\linewidth]{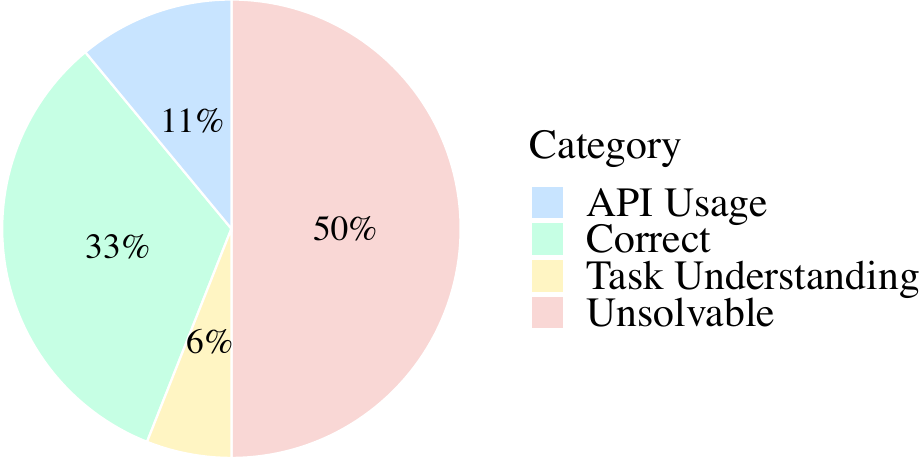} 
    \vspace{-6mm}
    \caption{Error analysis on 100 WebArena tasks.}
    \label{fig:analysis}
    \vspace{-5mm}
\end{figure}

\subsection{Error Analysis} We randomly sampled 100 tasks from WebArena and performed error analysis on the API-Based Agent. Figure \ref{fig:analysis} shows the distribution of error categories among these 100 tasks. We found that 33\% of the tasks are correctly performed with only API calling, 50\% are unsolvable with solely APIs, 6\% are incorrect due to incorrect task understanding, and 11\% are incorrect due to error in calling APIs such as mal-formatting and wrong input. In other words, among the 50 API solvable tasks, 66\% are performed correctly by the API-Based Agent. This showcases the strong capability of the API-Based Agent when given sufficient APIs to solve the task.

Additionally, the average API calls required to solve API solvable tasks are 2.1 API calls, demonstrating how API calling could reduce operational complexity for web tasks. Although the API-Based Agent took an average of 7.8 steps to complete WebArena tasks, most of the steps were taken to retrieve API documentation, resolve errors from it's previous generations, or verify it's outputs.

\subsection{Case Studies}
\label{appendix:case}
\begin{figure}[!h]
  \begin{center}
    \includegraphics[width=\linewidth]{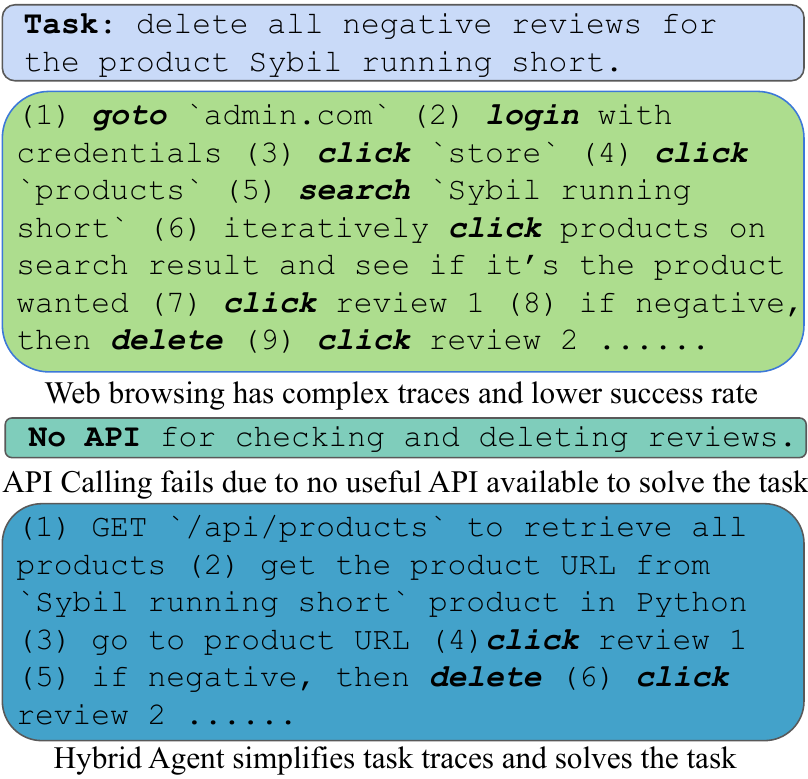}
  \end{center}
  \caption{The \hlhybrid{Hybrid Agent} succeeds while the \hlbrowse{Browsing Agent} and \hlapi{API-Based Agent} both fail}
  \label{fig:case1}
  \vspace{-5mm}
\end{figure}

\begin{figure}[!h]
  \begin{center}
\includegraphics[width=\linewidth]{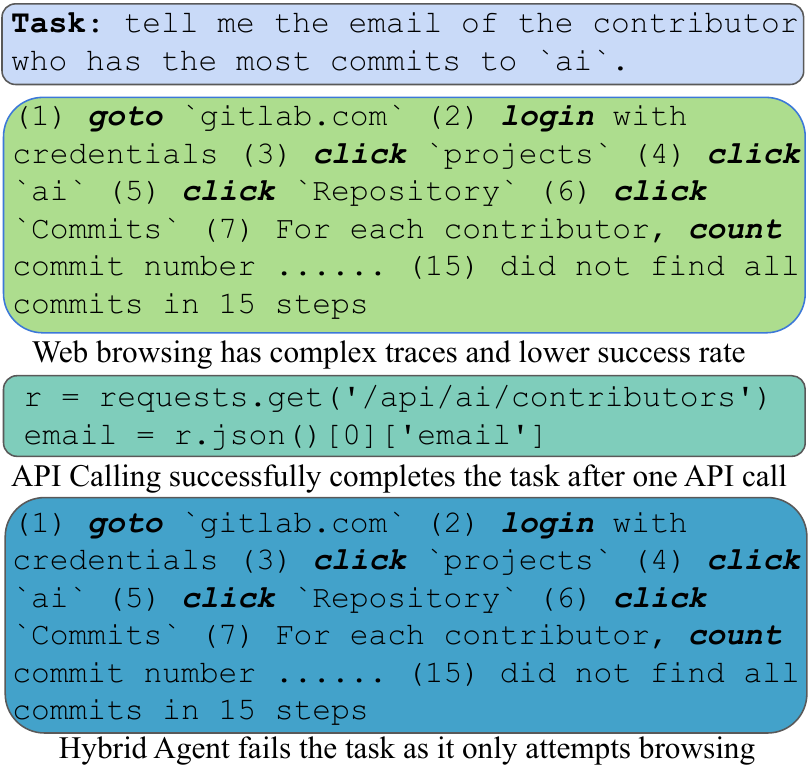}
  \end{center}
  \caption{The \hlapi{API-Based Agent} succeeds while the \hlbrowse{Browsing Agent} and the \hlhybrid{Hybrid Agent} fails.}
  \label{fig:case2}
\end{figure}

We analyze two contrasting instances as shown in Figure \ref{fig:case1} and Figure \ref{fig:case2}, where the Hybrid Agent and API-Based Agent exhibited different levels of performance on WebArena tasks. These case studies highlight the strengths and weaknesses of each agent, demonstrating scenarios where hybrid browsing outperforms API-only or browsing-only approaches, as well as cases where the API-Based Agent excels over the hybrid method.

\paragraph{Case 1} One example where the Hybrid Agent succeeded, while both the API-Based and Browsing Agents failed, involved a task from the Shopping Admin domain. The query was to ``delete all negative reviews for Sybil running short'', a product listed in the shopping admin interface. In this instance, the API-Based Agent failed because no relevant API endpoints were available for retrieving or deleting reviews. Similarly, the Browsing Agent failed, as completing this task purely through web navigation required too many steps, as depicted in Figure \ref{fig:case1}. This complexity made the task challenging for an agent relying solely on web interactions.
However, the Hybrid Agent successfully completed the task by leveraging both API and browsing functionalities. An example trace of the Hybrid Agent shown in Figure \ref{fig:case1}.
This case highlights the Hybrid Agent's ability to efficiently combine API calls with web interactions, allowing it to tackle complex multi-step tasks that would be difficult or impossible for solely browsing or solely API-Based Agents.

\paragraph{Case 2}

Conversely, there are instances where the API-Based Agent outperforms the Hybrid Agent. 
One such case occurred in the GitLab website, where the task was to \texttt{"}tell me the email address of the contributor who has the most commits to \texttt{ai}.\texttt{"} The API-Based Agent successfully completed this task by utilizing the \texttt{GET /api/{id}/contributors} API endpoint to retrieve the contributor with the highest number of commits and their associated email address.
On the other hand, the Hybrid Agent attempted to solve the task through browsing but encountered significant challenges. Accessing this information through web browsing required navigating GitLab's interface, locating the correct repository and branch, and identifying the top contributor manually, a task that might be too difficult to perform through web navigation alone. As a result, both the Browsing Agent and the Hybrid Agent failed to complete the task. This case demonstrates an example where API access provides a more straightforward solution than browsing in contexts requiring structured data retrieval.

\section{Conclusion}
In this paper, we propose new web agents that use APIs instead of traditional browsers. We find that API-Based Agents outperform Browsing Agents, especially on websites with good API support. Thus we further propose Hybrid Agents capable of interleaving API calling and browsing that empirically outperforms agents that only use one of the two interfaces. 


\section{Limitations}

\paragraph{API Availability} A key limitation of API-Based Agents is the inconsistent availability and coverage of APIs across websites. Even platforms with extensive API ecosystems, such as GitLab, may lack support for specific functionalities (e.g., retrieving a user's official username from a displayed name), leading to edge cases where API-Based Agents are unable to complete tasks due to incomplete API support. However, advancements in techniques like Automatic Web API Mining (AWM) \cite{wang2024agent} could potentially address this limitation by automatically generating APIs for unsupported tasks, reducing reliance on manual API creation.

\paragraph{Incorporating APIs} Unlike Browsing Agents, which can adapt to new websites without manual intervention, the API-Based Agent requires additional effort to integrate the necessary APIs documentation to the action space of the agent for each website. This manual integration process increases complexity, particularly when the agent must support a wide range of websites, limiting scalability compared to agents that rely solely on web browsing for interactions. However, future advancements could explore automatically inducing APIs using methods such as Agent Workflow Memory (AWM) \citep{wang2024agent} and self-improving \citep{zheng2025skillweaver}. These methods could identify and generate API calls for websites lacking formal API support, further expanding the applicability and efficiency of API-Based Agents. By automating the discovery and utilization of APIs, we envision even more robust agents capable of handling diverse web tasks without reliance on interaction through browsing.

\paragraph{Evaluation Benchmark} In this paper, we evaluate web agents exclusively on WebArena tasks. While WebArena offers realistic and diverse challenges, the number and variety of tasks may be limited. Other benchmarks, such as Webshop \citep{yao2022webshop}, MiniWoB \citep{pmlr-v70-shi17a}, Mind2Web \citep{deng2023mind2web}, WebVoyager \citep{he2024webvoyager}, and VisualWebArena \citep{koh-etal-2024-visualwebarena}, provide alternative valuable evaluation platforms. However, as discussed in Section \ref{subsec:benchmark}, WebArena aligns more closely with real-world scenarios and our use case, while other benchmarks lack support for API calling. For example, VisualWebArena is less applicable to our study because WebArena APIs lack support for interacting with images, a core component of VisualWebArena tasks. This could be potentially solved by aforementioned future approaches to automatically induce APIs that support image interactions. Nevertheless, we consider our work as a foundation for such future explorations. 

\bibliography{custom}

\appendix
\section{Appendix}

\subsection{Related Work}

The development of AI agents that interact with the web and APIs has garnered significant research attention. Web browsers, serving as the primary interface for interacting with online content, have long been a focus for AI research. Web-based agents that can navigate websites, extract information, and perform tasks autonomously have been studied extensively, especially in the context of LLMs and agents designed to mimic human behavior online.


\paragraph{Web Navigation Agents}



Much prior work has centered around agents that perform web-based tasks using browsing actions \citep{yao2022webshop,lai2024autowebglm,koh2024tree,pan2024autonomous}. These agents are particularly effective in environments where human-like interaction with a user interface is necessary \citep{workarena2024}. Frameworks such as WebArena have further refined the evaluation of such agents by providing complex and realistic web navigation tasks \citep{zhou2023webarena}. Our work explores the Hybrid Agent that combines web browsing with API interactions. While prior work primarily focuses on browsing-only agents, we examine how Hybrid Agents can enhance performance by integrating structured API calls with web navigation.

\paragraph{Code Generation Agents and Tool Usage}
Another stream of research focuses on agents that interact with online content via application programming interfaces (APIs) \citep{wang2022what,patil2023gorilla,qin2023toolllm,yuan2024easytoolenhancingllmbasedagents,wang2024opendevinopenplatformai,du2024anytool}. In this context, works such as CodeAct have pioneered the development of agents that generate and execute code, including API calls, to perform tasks typically reserved for software engineers \citep{wang2024executable, zhang2024codeagentenhancingcodegeneration, tang2024codeagentautonomouscommunicativeagents}. These API-Based Agents are optimized for tasks that involve structured data exchanges, allowing them to perform operations more efficiently than traditional web navigation agents \citep{shen2024shortcutsbenchlargescalerealworldbenchmark}. On the other hand, our work integrates both browsing and API interactions, demonstrating that Hybrid Agents can outperform API-only agents in tasks requiring web navigation. While existing research shows the efficiency of API-Based Agents, our Hybrid Agent dynamically switches between APIs and web browsing to optimize task performance.

Additionally, we are the first to explore comparative studies of API v.s. Browsing Agents on the same websites. We demonstrate that API-Based Agents are often more efficient than Browsing Agents when APIs are available, leading to significant improvements in performance. 
This finding is aligned with previous studies that highlight the advantages of structured interactions through APIs compared to unstructured web browsing interactions.

\subsection{WebArena Tasks}

WebArena reproduces the functionality of several commonly-used websites using open-source frameworks, with real-world data imported into the reproduced websites.

WebArena includes tasks related to the following websites:

\label{appendix:webarena}
\begin{itemize} 
    \item \textbf{Gitlab}\footnote{Original Website: \url{https://gitlab.com}} – 180 instances: This website contains tasks related to project management and version control, where agents perform tasks like opening issues, handling merge requests, or creating repositories. Example query: Submit a merge request for \texttt{a11yproject.com/redesign} branch to be merged into the \texttt{markdown-figure-block} branch, assign myself as the reviewer.
    
    \item \textbf{Map}\footnote{Original Website: \url{https://www.openstreetmap.org}} – 109 instances: For this website, tasks are centered around navigation, trip planning and queries about distances, requiring the agent to retrieve and interpret map-based data, similar to using real-world map services like Google map. Example query: Tell me the full address of all international airports that are within a driving distance of 50 km to Carnegie Mellon University.
    
    \item \textbf{Shopping}\footnote{Developed using Adobe Magento (\url{https://github.com/magento/magento2})} – 187 instances: Tasks related to this website represents typical e-commerce tasks, such as searching for products, adding items to carts, and processing transactions. Example query: Change the delivery address for my most recent order to 77 Massachusetts Ave, Cambridge, MA.
    
    \item \textbf{Shopping Admin}\footnote{Developed using Adobe Magento (\url{https://github.com/magento/magento2})} – 182 instances: This setting involves managing backend administrative tasks for an online store, like managing product inventories, processing orders, or viewing sales reports. Example query: Tell me the the number of reviews that our store received by far that mention term ``\texttt{satisfied}''.
    
    \item \textbf{Reddit}\footnote{Deployed Postmill (\url{https://postmill.xyz/}), the open-sourced counterpart of Reddit (\url{https://www.reddit.com})} – 106 instances: Tasks here are similar to interactions with the official Reddit, where agents need to post comments, upvote or down-vote posts, or retrieve information from threads. Example query: Tell me the count of comments that have received more downvotes than upvotes for the user who made the latest post on the Showerthoughts forum.
    
    \item \textbf{Multi-Website Tasks} – 48 instances: These examples involve tasks that span across two websites, requiring the agent to interact with both websites simultaneously, adding complexity to the task. Example query: Create a folder named news in gimmiethat.space repo. Within it, create a file named urls.txt that contains the URLs of the 5 most recent posts from the news related subreddits?
\end{itemize}

\subsection{Obtaining APIs of WebArena Websites}
\label{appendix:apis}
\begin{itemize}
    \item \textbf{Gitlab}: we leveraged the open Gitlab REST APIs\footnote{Documentation of all Gitlab APIs could be found at \url{https://docs.gitlab.com/ee/api/rest/}.}, consisting of 988 endpoints. Most of WebArena tasks are covered by these APIs, with only a small fraction of tasks, such as retrieving users' Gitlab feed token, are not covered by any existing endpoints,
    \item \textbf{Map}: The Map website offers three sets of APIs, each offering distinct functionalities, with a total of 53 endpoints. The first set of APIs, openly available at Nominatim\footnote{The API documentations could be found at \url{https://nominatim.org/release-docs/develop/api/Overview/}}, offers essential endpoints for geographic searches. The second set of APIs, from Project OSRM\footnote{Documentations of APIs available at \url{https://project-osrm.org/docs/v5.5.1/api}}, focuses on routing and navigation functionalities. The third set of APIs, available at OpenStreetMap\footnote{API documentations openly available at \url{https://wiki.openstreetmap.org/wiki/API_v0.6}}, deals primarily with map database operations. This API is rarely used in WebArena tasks but offers capabilities for interacting with OSM data. 
    \item \textbf{Shopping}: The e-commerce website uses APIs from the Adobe Commerce API\footnote{APIs documented at \url{https://developer.adobe.com/commerce/webapi/rest/quick-reference/}}, consisting of 556 endpoints. These endpoints provide support for common shopping tasks such as purchasing products, searching categories, and managing customer accounts.
    \item \textbf{Shopping Admin}: This website shares a common set of APIs with the shopping website. However, this website requires a unique admin token to access the admin-only APIs such as changing the price of products and deleting products from stores.
    \item \textbf{Reddit}: The Reddit tasks in WebArena are based on a self-hosted limited clone of the Reddit website~\footnote{\url{https://codeberg.org/Postmill/Postmill}}, with limited functionalities as compared to the official site. As a result, all of the available APIs are self-implemented, with a best effort to mimic to official Reddit APIs. This website supports 31 endpoints, which include writing comments and voting posts.
    \item \textbf{Multi-Website Tasks}: we provide APIs from all websites included in the task to the agents, where we explicitly state which set of APIs belongs to which website. This could allow agents to identify the correct set of APIs to use when transitioning between websites. It is also worth noting that the framework of our agents supports a unified workspace that allows the agents to carry over the information from one website to another.
\end{itemize}

\subsection{Additional Analysis}

\label{appendix:analysis}

Table \ref{task_percent} documents the percentage of actions of our Hybrid Agent. Across all websites, our Hybrid Agent chooses to do both Browsing and API in the same task at least half of the time. 

\begin{table*}[t]
\centering

\begin{tabular}{l|rrrrrr|r}
\toprule
\textbf{Actions} & \textbf{Gitlab} & \textbf{Map} & \textbf{Shopping} & \textbf{Admin} & \textbf{Reddit} & \textbf{Multi} & \textbf{AVG.} \\ \midrule
Browsing only & 7.8 & 3.7 & 38.5 & 2.2 & 0 & 8.3 & 12.1 \\
API only & 21.1 & 4.6 & 7.5 & 1.1 & 0 & 10.4 & 7.9 \\
Browsing+API & 71.1 & 91.7 & 54.0 & 96.7 & 82.1 & 81.3 & 80.0 \\ 
\bottomrule
\end{tabular}
\caption{Percentage of Actions (\%) that our Hybrid Agent takes for each type of tasks. Each column sums up to 1. }
\label{task_percent}
\end{table*}

\begin{table*}[t]
\centering
\resizebox{\linewidth}{!}{
\begin{tabular}{l|rrrrrr|r}
\toprule
\textbf{Choices of Action} & \textbf{Gitlab} & \textbf{Map} & \textbf{Shopping} & \textbf{Admin} & \textbf{Reddit} & \textbf{Multi} & \textbf{AVG.} \\ \midrule
Browsing only & 7.1(1/14) & 50.0(2/4) & 23.6(17/72) & 50.0(2/4) & 0(0/0) & 25.0(1/4) & 23.5(23/98) \\
API only & 47.4(18/38) & 40.0(2/5) & 21.4(3/14) & 50.0(1/2) & 0.0(0/0) &  20.0(1/5) & 39.1(25/64) \\
Browsing+API & 47.7(61/128) & 46.0(46/100) & 27.7(28/101) & 40.9(72/176) & 51.9(55/106) & 15.4(6/39) & 41.2(268/650) \\ 
\bottomrule
\end{tabular}}
\caption{The accuracy (\%) of the Hybrid Agent across choices of actions for each website, with the number of correct instances / number of total instances in parentheses.
}
\label{task_percent_acc}
\end{table*}

Table \ref{task_percent_acc} documents the accuracy of the Hybrid Agent across websites when performing different choices of actions. It shows consistently high accuracy when choosing API only and API+browsing.

Table \ref{tab:steps} shows the breakdown of number of steps and cost by website.

\begin{table*}[ht]
\centering
\resizebox{\textwidth}{!}{%
\begin{tabular}{l|rrrrrrrrrrrrrr}
\toprule
\multirow{2}{*}{\textbf{Agents}} & 
\multicolumn{2}{c}{\textbf{Gitlab}} & 
\multicolumn{2}{c}{\textbf{Map}} & \multicolumn{2}{c}{\textbf{Shopping}} & \multicolumn{2}{c}{\textbf{Shop-Admin}} & \multicolumn{2}{c}{\textbf{Reddit}} & \multicolumn{2}{c}{\textbf{Multi Sites}} & \multicolumn{2}{c}{\textbf{AVG.}} \\ \cmidrule(l){2-15} 
                    & steps & cost & steps & cost & steps & cost & steps & cost & steps & cost & steps & cost & steps & cost \\ \midrule 
Browsing & 9.4 & 0.2 & 8.0 & 0.1 & 7.3 & 0.1 & 7.0 & 0.2 & 11.1 & 0.1 & 7.5 & 0.1 & 8.4 & 0.1 \\
API-Based & 7.0 & 1.7 & 6.6 & 1.1 & 8.2 & 1.0 & 8.4 & 1.1 & 8.8 & 0.6 & 7.7 & 1.6 & 7.8 & 1.2  \\
Hybrid & 8.1 & 2.0 & 9.4 & 1.7 & 8.2 & 1.3 & 9.0 & 1.4 & 7.8 & 0.6 & 8.0 & 1.9 & 8.5 & 1.4
\\ \bottomrule
\end{tabular}%
}
\caption{Number of Steps and Cost (in U.S. dollars) of Agents across WebArena Websites}
\label{tab:steps}
\end{table*}

\begin{figure}[!h]
    \centering
    \includegraphics[width=\linewidth]{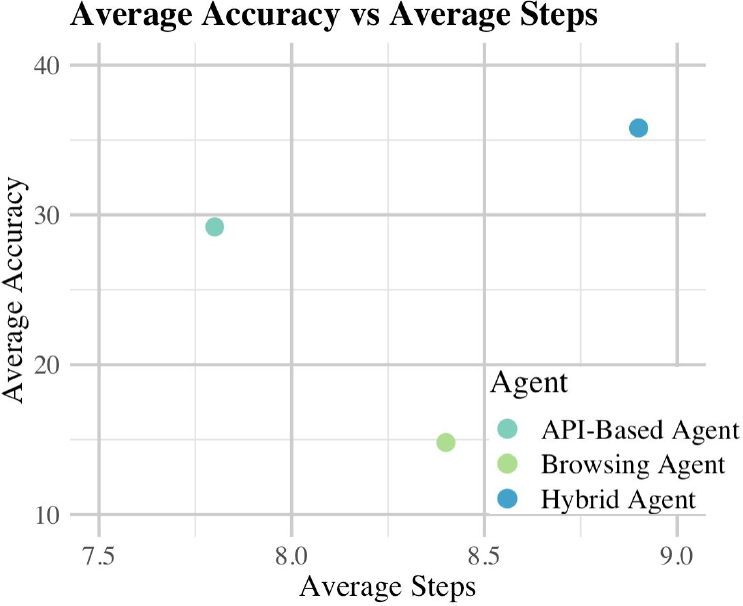} 
    \caption{Steps of agents on WebArena.}
    \label{fig:step}
\end{figure}

\paragraph{Steps} Figure \ref{fig:step} demonstrates a scatterplot of the average accuracy of each agent on WebArena over their average steps. The Browsing Agent takes more steps to complete tasks compared to the API-Based Agent on average, while the Hybrid Agent takes the most steps amongst the three agents. This is likely due to the Browsing Agent's reliance on navigating web interfaces and interacting with visual elements, which involves a sequential and more time consuming processes. The API-Based Agent is the most efficient in terms of steps, as it can directly interact with structured data via APIs, bypassing many of the steps involved in traditional web navigation. The Hybrid Agent, combining both action spaces from the Browsing Agent and the API-Based Agent, takes more steps than both agents. 

\begin{figure}[!h]
    \centering
    \includegraphics[width=\linewidth]{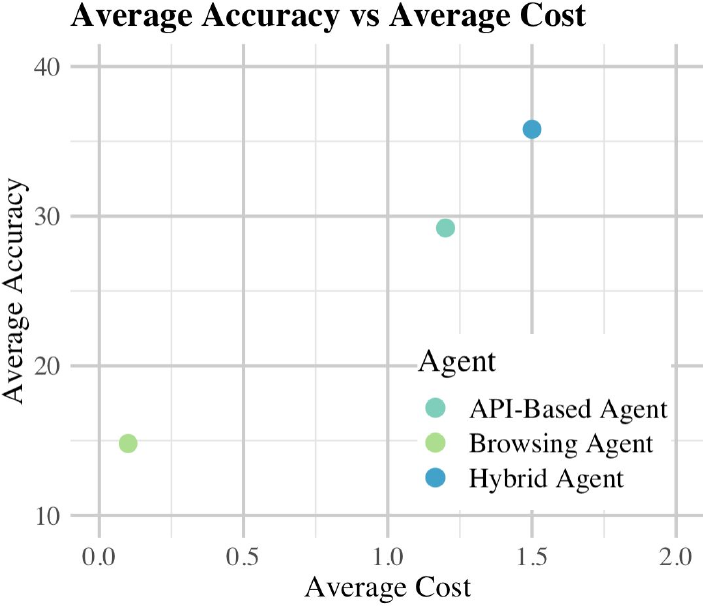} 
    \caption{Costs of agents on WebArena.}
    \label{fig:cost}
\end{figure}
\paragraph{Costs} Figure \ref{fig:cost} demonstrates a scatterplot of the average accuracy of each agent on WebArena over their average costs. The cost of completing tasks shows a different trend. While the Browsing Agent requires more steps, it is much cheaper compared to the API-Based Agent and the Hybrid Agent. This is primarily because the prompts needed for Browsing Agents are much shorter. When browsing, the agent only needs instructions on how to use the web interface and the limited action space around 14 browsing actions. In contrast, API-Based and Hybrid Agents require access to a much larger set of API calls. For example, when interacting with GitLab, the agent is provided with 988 available APIs, leading to much longer prompts and significantly increasing the cost of execution. The cost goes down when the prompt for API calling is shorter. For example, the Reddit website has the least length of API documentation, where its cost is also less than other websites. However, as visualized in Figure \ref{fig:cost}, the accuracy of the API-Based Agent and the Hybrid Agent is much higher than the Browsing Agent, which makes the increase in cost justifiable due to the significantly improved task performance. The higher cost is offset by the agents' ability to complete tasks more accurately and efficiently. In the future, this increased cost could potentially be mitigated by methods that retrieve only relevant APIs on the fly.

\subsection{API-Based Agent Prompt}
\label{appendix:api-prompt}

\begin{promptbox}[Full System Prompt]{api}
\texttt{Full System Prompt} = \texttt{System Prefix} + \texttt{API Prompt} + \texttt{System Suffix}
\end{promptbox}

\begin{promptbox}[System Prefix]{api}
You are an AI assistant that performs tasks on the websites. You should give helpful, detailed, and polite responses to the user's queries.

You have the ability to call site-specific APIs using Python, or browse the website directly. 

\end{promptbox}

\begin{promptbox}[API Prompt]{api}
To call APIs, you can use an interactive Python (Jupyter Notebook) environment, executing code with \texttt{<execute\_ipython>}.

\texttt{<execute\_ipython>}\\
\texttt{print(``Hello World'')}\\
\texttt{</execute\_ipython>}

This can be used to call the Python requests library, which is already installed for you. Here are some hints about effective API usage:
\begin{itemize}
    \item It is better to actually view the API response and ensure the relevant information is correctly extracted and utilized before attempting any programmatic parsing.
    \item Make use of HTTP headers when making API calls, and be careful of the input parameters to each API call.
    \item Be careful about pagination of the API response, the response might only contain the first few instances, so make sure you look at all instances.
\end{itemize}

The user will provide you with a list of API calls that you can use.

\end{promptbox}

\begin{promptbox}[System Suffix]{api}
The information provided by the user might be incomplete or ambiguous. For example, if I want to search for \texttt{``xyz''}, then \texttt{``xyz''} could be the name of a product, a user, or a category on the site. In these cases, you should attempt to evaluate all potential cases that the user might be indicating and be careful about nuances in the user's query. Also, do NOT ask the user for any clarification, they cannot clarify anything and you need to do it yourself.

When you think you successfully finished the task, first respond with \texttt{Finish[answer]} where you include \textit{only} your answer to the question \texttt{[]} if the user asks for an answer, make sure you should only include the answer to the question but not any additional explanation, details, or commentary unless specifically requested.

After that, when you responded with your answer, you should respond with \texttt{<finish></finish>}.

Then finally, to exit, you can run\\
\texttt{<execute\_bash>}\\
\texttt{exit()}\\
\texttt{</execute\_bash>}

Your responses should be concise.
The assistant should attempt fewer things at a time instead of putting too many commands OR too much code in one \texttt{execute} block.

Include AT MOST ONE \texttt{<execute\_ipython>}, \texttt{<execute\_browse>}, or \texttt{<execute\_bash>} per response.

IMPORTANT: Execute code using \texttt{<execute\_ipython>}, \texttt{<execute\_bash>}, or \texttt{<execute\_browse>} whenever possible.

Below are some examples:\\
--- START OF EXAMPLE ---\\
\texttt{Examples}\\
--- END OF EXAMPLE ---\\
Now, let's start!

\end{promptbox}

\begin{promptbox}[Initial User Prompt]{api}

Think step by step to perform the following task related to gitlab. Answer the question: ***\texttt{Example WebArena Intent}***

The site URL is \texttt{Example Site URL}, use this instead of the normal site URL. 

For API calling, use this access token: \texttt{Example Access Token}.

My username on this website is \texttt{Example Username}.

Below is the list of all APIs you can use and their descriptions:

\texttt{Example API Documentation.}

Note: Before actually using a API call, *you should call the \texttt{get\_api\_documentation} function in the \texttt{utils} module to get detailed API documentation of the API.* For example, if you want to use the API \texttt{GET /api/v4/projects/{id}/repository/commits}, you should first do: 

\texttt{<execute\_ipython>}\\
\texttt{from utils import get\_api\_documentation}\\
\texttt{get\_api\_documentation(``GET /api/v4/projects/\{id\}/repository/commits'')}\\
\texttt{</execute\_ipython>}

This will provide you with detailed descriptions of the input parameters and example output jsons.

\end{promptbox}

\subsection{Hybrid Agent Prompt}
\label{appendix:hybrid-prompt}

\begin{promptbox}[Full System Prompt]{hybrid}
\texttt{Full System Prompt} = \texttt{System Prefix} + \texttt{API Prompt} + \texttt{Browsing Prompt} + \texttt{System Suffix}
\end{promptbox}

\begin{promptbox}[System Prefix]{hybrid}
You are an AI assistant that performs tasks on the websites. You should give helpful, detailed, and polite responses to the user's queries.

You have the ability to call site-specific APIs using Python, or browse the website directly. 

IMPORTANT: In general, you must always first try to use APIs to perform the task; however, you should use web browsing when there is no useful API available for the task. 

IMPORTANT: After you tried out using APIs, you must use web browsing to navigate to some URL containing contents that could verify whether the results you obtained by API calling is correct. 

\end{promptbox}

\begin{promptbox}[API Prompt]{hybrid}
To call APIs, you can use an interactive Python (Jupyter Notebook) environment, executing code with \texttt{<execute\_ipython>}.

\texttt{<execute\_ipython>}\\
\texttt{print(``Hello World!'')}\\
\texttt{</execute\_ipython>}

This can be used to call the Python requests library, which is already installed for you. Here are some hints about effective API usage:
\begin{itemize}
    \item It is better to actually view the API response and ensure the relevant information is correctly extracted and utilized before attempting any programmatic parsing.
    \item Make use of HTTP headers when making API calls, and be careful of the input parameters to each API call.
    \item Be careful about pagination of the API response, the response might only contain the first few instances, so make sure you look at all instances.
\end{itemize}

The user will provide you with a list of API calls that you can use.

\end{promptbox}

\begin{promptbox}[Browsing Prompt]{hybrid}

You can browse the Internet by putting special browsing commands within \texttt{<execute\_browse>} and \texttt{</execute\_browse>} (in Python syntax).

For example to select the option \texttt{blue} from the dropdown menu with bid \texttt{12}, and click on the submit button with bid \texttt{51}:

\texttt{<execute\_browse>}\\
\texttt{select\_option(``12'', ``blue'')}\\
\texttt{click(``51'')}\\
\texttt{</execute\_browse>}\\

The following actions are available:\\

\texttt{def goto(url: str):} \\
\hspace*{1em}\texttt{``````Navigate to the specified URL.} \\
\hspace*{1em}\texttt{Examples:} \\
\hspace*{2em}\texttt{goto(``http://www.example.com'')} \\
\hspace*{1em}\texttt{''''''}
\hspace*{1em}\texttt{} \\

\texttt{def go\_back():} \\
\hspace*{1em}\texttt{``````Navigate back to the previous page.} \\
\hspace*{1em}\texttt{Examples:} \\
\hspace*{2em}\texttt{go\_back()} \\
\hspace*{1em}\texttt{''''''}
\hspace*{1em}\texttt{} \\

\texttt{def go\_forward():} \\
\hspace*{1em}\texttt{``````Navigate forward to the next page.} \\
\hspace*{1em}\texttt{Examples:} \\
\hspace*{2em}\texttt{go\_forward()} \\
\hspace*{1em}\texttt{''''''}
\hspace*{1em}\texttt{} \\

\texttt{def scroll(delta\_x: float, delta\_y: float):} \\
\hspace*{1em}\texttt{``````Scroll the page by the specified amount.} \\
\hspace*{1em}\texttt{Examples:} \\
\hspace*{2em}\texttt{scroll(0, 200)} \\
\hspace*{2em}\texttt{scroll(-50.2, -100.5)} \\
\hspace*{1em}\texttt{''''''}
\hspace*{1em}\texttt{} \\

\texttt{def fill(bid: str, value: str):} \\
\hspace*{1em}\texttt{``````Fill the input field with the specified value.} \\
\hspace*{1em}\texttt{Examples:} \\
\hspace*{2em}\texttt{fill(``237'', ``example value'')} \\
\hspace*{2em}\texttt{fill(``45'', ``multi-line\ example'')} \\
\hspace*{2em}\texttt{fill(``a12'', ``example with ``quotes'''')} \\
\hspace*{1em}\texttt{''''''}
\hspace*{1em}\texttt{} \\

\texttt{def select\_option(bid: str, options: str | list[str]):} \\
\hspace*{1em}\texttt{``````Select an option from a dropdown menu.} \\
\hspace*{1em}\texttt{Examples:} \\
\hspace*{2em}\texttt{select\_option(``48'', ``blue'')} \\
\hspace*{2em}\texttt{select\_option(``48'', [``red'', ``green'', ``blue''])} \\
\hspace*{1em}\texttt{''''''}
\hspace*{1em}\texttt{} \\

\texttt{def focus(bid: str):} \\
\hspace*{1em}\texttt{``````Focus on an element.} \\
\hspace*{1em}\texttt{Examples:} \\
\hspace*{2em}\texttt{focus(``b455'')} \\
\hspace*{1em}\texttt{''''''}
\hspace*{1em}\texttt{} \\
\end{promptbox}

\begin{promptbox}[Browsing Prompt - Continued]{hybrid}
\texttt{def click(bid: str, button: Literal[``left'', ``middle'', ``right''] = ``left'', modifiers: list[typing.Literal[``Alt'', ``Control'', ``Meta'', ``Shift'']] = []):} \\
\hspace*{1em}\texttt{``````Click on an element with the specified button and modifiers.} \\
\hspace*{1em}\texttt{Examples:} \\
\hspace*{2em}\texttt{click(``51'')} \\
\hspace*{2em}\texttt{click(``b22'', button=``right'')} \\
\hspace*{2em}\texttt{click(``48'', button=``middle'', modifiers=[``Shift''])} \\
\hspace*{1em}\texttt{''''''}
\hspace*{1em}\texttt{} \\

\texttt{def dblclick(bid: str, button: Literal[``left'', ``middle'', ``right''] = ``left'', modifiers: list[typing.Literal[``Alt'', ``Control'', ``Meta'', ``Shift'']] = []):} \\
\hspace*{1em}\texttt{``````Double-click on an element with the specified button and modifiers.} \\
\hspace*{1em}\texttt{Examples:} \\
\hspace*{2em}\texttt{dblclick(``12'')} \\
\hspace*{2em}\texttt{dblclick(``ca42'', button=``right'')} \\
\hspace*{2em}\texttt{dblclick(``178'', button=``middle'', modifiers=[``Shift''])} \\
\hspace*{1em}\texttt{''''''}
\hspace*{1em}\texttt{} \\

\texttt{def hover(bid: str):} \\
\hspace*{1em}\texttt{``````Hover over an element.} \\
\hspace*{1em}\texttt{Examples:} \\
\hspace*{2em}\texttt{hover(``b8'')} \\
\hspace*{1em}\texttt{''''''}
\hspace*{1em}\texttt{} \\

\texttt{def press(bid: str, key\_comb: str):} \\
\hspace*{1em}\texttt{``````Press a key combination on an element.} \\
\hspace*{1em}\texttt{Examples:} \\
\hspace*{2em}\texttt{press(``88'', "Backspace")} \\
\hspace*{2em}\texttt{press(``a26'', ``Control+a'')} \\
\hspace*{2em}\texttt{press(``a61'', ``Meta+Shift+t'')} \\
\hspace*{1em}\texttt{''''''}
\hspace*{1em}\texttt{} \\

\texttt{def clear(bid: str):} \\
\hspace*{1em}\texttt{``````Clear the input field.} \\
\hspace*{1em}\texttt{Examples:} \\
\hspace*{2em}\texttt{clear(``996'')} \\
\hspace*{1em}\texttt{''''''}
\hspace*{1em}\texttt{} \\

\texttt{def drag\_and\_drop(from\_bid: str, to\_bid: str):} \\
\hspace*{1em}\texttt{``````Drag and drop an element to another element.} \\
\hspace*{1em}\texttt{Examples:} \\
\hspace*{2em}\texttt{drag\_and\_drop(``56'', ``498'')} \\
\hspace*{1em}\texttt{''''''}
\hspace*{1em}\texttt{} \\

\texttt{def upload\_file(bid: str, file: str | list[str]):} \\
\hspace*{1em}\texttt{``````Upload a file to the specified element.} \\
\hspace*{1em}\texttt{Examples:} \\
\hspace*{2em}\texttt{upload\_file(``572'', ``my\_receipt.pdf'')} \\
\hspace*{2em}\texttt{upload\_file(``63'', [``/home/bob/Documents/image.jpg'', ``/home/bob/Documents/file.zip''])} \\
\hspace*{1em}\texttt{''''''}
\hspace*{1em}\texttt{} \\

\end{promptbox}

\begin{promptbox}[System Suffix]{hybrid}
The information provided by the user might be incomplete or ambiguous. For example, if I want to search for \texttt{``xyz''}, then \texttt{``xyz''} could be the name of a product, a user, or a category on the site. In these cases, you should attempt to evaluate all potential cases that the user might be indicating and be careful about nuances in the user's query. Also, do NOT ask the user for any clarification, they cannot clarify anything and you need to do it yourself.

When you think you successfully finished the task, first respond with \texttt{Finish[answer]} where you include \textit{only} your answer to the question \texttt{[]} if the user asks for an answer, make sure you should only include the answer to the question but not any additional explanation, details, or commentary unless specifically requested.

After that, when you responded with your answer, you should respond with \texttt{<finish></finish>}.

Then finally, to exit, you can run\\
\texttt{<execute\_bash>}\\
\texttt{exit()}\\
\texttt{</execute\_bash>}

Your responses should be concise.
The assistant should attempt fewer things at a time instead of putting too many commands OR too much code in one \texttt{execute} block.

Include AT MOST ONE \texttt{<execute\_ipython>}, \texttt{<execute\_browse>}, or \texttt{<execute\_bash>} per response.

IMPORTANT: Execute code using \texttt{<execute\_ipython>}, \texttt{<execute\_bash>}, or \texttt{<execute\_browse>} whenever possible.

Below are some examples:\\
--- START OF EXAMPLE ---\\
\texttt{Examples}\\
--- END OF EXAMPLE ---\\
Now, let's start!

\end{promptbox}

\begin{promptbox}[Initial User Prompt]{hybrid}

Think step by step to perform the following task related to gitlab. Answer the question: ***\texttt{Example WebArena Intent}***

The site URL is \texttt{Example Site URL}, use this instead of the normal site URL. 

For API calling, use this access token: \texttt{Example Access Token}.

For web browsing, You should start from the URL \texttt{Example Start URL}, and this webpage is already logged in and opened for you. 

My username on this website is \texttt{Example Username}.

Below is the list of all APIs you can use and their descriptions:

\texttt{Example API Documentation.}

Note: Before actually using a API call, *you should call the \texttt{get\_api\_documentation} function in the \texttt{utils} module to get detailed API documentation of the API.* For example, if you want to use the API \texttt{GET /api/v4/projects/{id}/repository/commits}, you should first do: 

\texttt{<execute\_ipython>}\\
\texttt{from utils import get\_api\_documentation}\\
\texttt{get\_api\_documentation(``GET /api/v4/projects/\{id\}/repository/commits'')}\\
\texttt{</execute\_ipython>}

This will provide you with detailed descriptions of the input parameters and example output jsons.

IMPORTANT: In general, you must always first try to use APIs to perform the task; however, you should use web browsing when there is no useful API available for the task. IMPORTANT: After you tried out using APIs, you must use web browsing to navigate to some URL containing contents that could verify whether the results you obtained by API calling is correct. 
\end{promptbox}

\end{document}